\pdfoutput=1

\documentclass[11pt]{article}

\usepackage[]{acl}

\usepackage{times}
\usepackage{latexsym}
\usepackage{amsmath}
\usepackage{tabularx}
\usepackage{multirow}
\usepackage{booktabs}
\usepackage{graphicx}
\usepackage[export]{adjustbox}
\usepackage{caption}
\usepackage{subcaption}
\usepackage{enumitem}
\usepackage{xspace}
\usepackage{scalerel,xparse}
\NewDocumentCommand\appleemoji{}{\scalerel*{\includegraphics{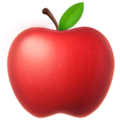}}{X}}

\usepackage[T1]{fontenc}

\usepackage[utf8]{inputenc}

\usepackage{microtype}

\usepackage{inconsolata}

\newcommand\testbed{\textsc{APPLS}\xspace}

\newcolumntype{L}[1]{>{\raggedright\let\newline\\\arraybackslash\hspace{0pt}}p{#1}}
\newcolumntype{M}[1]{>{\raggedright\let\newline\\\arraybackslash\hspace{0pt}}m{#1}}

\definecolor{darkgreen}{rgb}{0.0, 0.4, 0.13}

\title{\scalebox{1.4}{\appleemoji}~\testbed: Evaluating Evaluation Metrics for Plain Language Summarization}

\author{Yue Guo$^{1}$\thanks{Work performed while at University of Washington.} \quad Tal August$^{1}$ \quad Gondy Leroy$^{2}$ \quad Trevor Cohen$^{3}$ \quad Lucy Lu Wang$^{3,4}$ \\ [1mm]
  $^{1}$University of Illinois Urbana-Champaign
  \quad
  $^{2}$University of Arizona
  \\
  $^{3}$University of Washington \quad
  $^{4}$Allen Institute for AI  \\ [1mm]
  \texttt{\{yueg, taugust\}@illinois.edu; \{cohenta, lucylw\}@uw.edu} \\ [-1mm]
}

\begin{document}
\maketitle

\begin{abstract}
While there has been significant development of models for Plain Language Summarization (PLS), evaluation remains a challenge. PLS lacks a dedicated assessment metric, and the suitability of text generation evaluation metrics is unclear due to the unique transformations involved (e.g., adding background explanations, removing jargon). To address these questions, our study introduces a granular meta-evaluation testbed, \testbed, designed to evaluate metrics for PLS. We identify four PLS criteria from previous work---informativeness, simplification, coherence, and faithfulness---and define a set of perturbations corresponding to these criteria that sensitive metrics should be able to detect. We apply these perturbations to the texts of two PLS datasets to create our testbed. Using \testbed, we assess performance of 14 metrics, including automated scores, lexical features, and LLM prompt-based evaluations. Our analysis reveals that while some current metrics show sensitivity to specific criteria, no single method captures all four criteria simultaneously. We therefore recommend a suite of automated metrics be used to capture PLS quality along all relevant criteria. This work contributes the first meta-evaluation testbed for PLS and a comprehensive evaluation of existing metrics.\footnote{\testbed and our evaluation code 
can be found at \href{https://github.com/LinguisticAnomalies/APPLS}{https://github.com/LinguisticAnomalies/APPLS}.}
\end{abstract}


\section{Introduction}
\begin{figure}
    \centering
    \includegraphics[trim={20cm 14.2cm 32.7cm 13.2cm},clip,width=0.47\textwidth]{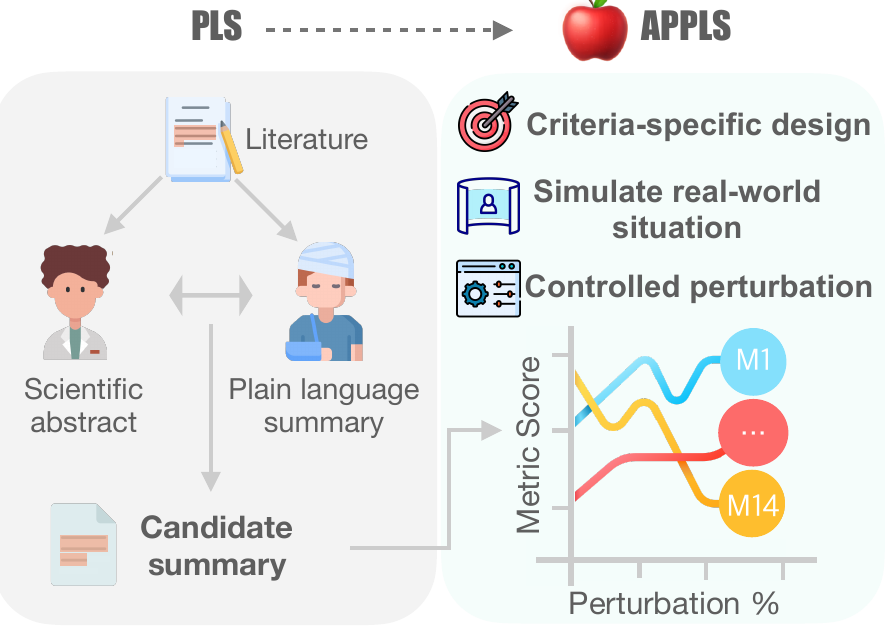}
    \caption{We present \testbed, the first granular testbed for analyzing evaluation metric performance for plain language summarization (PLS). We assess performance of 14 existing metrics, including automated scores, lexical features, and LLM prompt-based evaluations.
    }
    \label{fig:teaser_image}
\end{figure}

Plain language summaries of scientific information are important to make science more accessible \citep{kuehne2015lay, stoll2022plain} and inform public decision-making \citep{holmes2005evidence, pattisapu2020leveraging}. Recently, generative models have made gains in translating scientific information into plain language approachable to lay audiences \citep{august2022paper, goldsack2023domain, devaraj2021paragraph}. Despite these gains, the field has not reached consensus on effective automated evaluation metrics for plain language summarization (PLS) \cite{luo2022benchmarking, ondov2022survey} due to the multifaceted nature of the PLS task. Removal of unnecessary details \citep{pitcher2022template}, adding relevant background explanations \citep{guo2021automated}, jargon interpretation \citep{pitcher2022template}, and text simplification \citep{devaraj2021paragraph} are all involved in PLS, posing challenges for comprehensive evaluation. 

Our goal is to assess how well existing metrics capture the multiple criteria of PLS. We identify four criteria, informed by prior work \citep{pitcher2022template, ondov2022survey, stoll2022plain, jain2022survey}, that a PLS metric should be sensitive to: \textit{informativeness}, \textit{simplification}, \textit{coherence}, and \textit{faithfulness}. We introduce a set of perturbations to probe metric sensitivity to these criteria, where each perturbation is designed to affect a single criterion with ideally minimal impact to others.\footnote{We acknowledge that introducing any change in text likely affects multiple criteria, though we design our perturbations carefully to try and minimize these impacts.} By incrementally introducing perturbations to the texts of two scientific PLS datasets, CELLS \cite{guo2022cells} and PLABA \cite{attal2023dataset}, we create our meta-evaluation testbed \testbed. 

We analyze 14 metrics using \testbed, including the most widely used metrics in text simplification and summmarization literature, and recently-proposed prompt-based methods \citep{gao2023human, luo2023chatgpt}. We find that established metrics like ROUGE \citep{lin2004rouge}, BERTScore \citep{zhang2019bertscore}, and QAEval \citep{deutsch2021towards} do not capture simplification and are inconsistent at capturing perturbations in  
informativeness, coherence, and faithfulness; SARI score \citep{xu2016optimizing}, explicitly crafted for text simplification, is the only automated score that displays sensitivity towards simplification perturbations but not to other perturbations. LLM prompt-based evaluations are effective for assessing informativeness, faithfulness, and simplification, but not for coherence. Our analysis suggests that a single overall score cannot simultaneously respond to all four criteria.

Our main contributions are as follows:
\begin{itemize}[itemsep=0pt, topsep=2pt, leftmargin=10pt]
\item We present \testbed, the first granular testbed for analyzing evaluation metric performance for plain language summarization; the testbed is created by applying 11 perturbations along four dimensions to two scientific PLS datasets (\S\ref{criteria_specific_perturbation_design}, \ref{perturbation_implementation});
\item We conduct a thorough analysis of 14 existing evaluation metrics (including automated metrics, lexical features, and LLM prompting methods), demonstrating mixed effectiveness in evaluating informativeness, coherence, faithfulness, and simplification (\S\ref{existing_metrics}, \ref{results});
\item Based on our findings, we recommend an evaluation strategy for PLS that combines multiple automated metrics able to capture differences along all relevant dimensions. 
\end{itemize}

\section{Related Work}

\vspace{-1mm}
\paragraph{Limitations of Existing Metrics}
The primary approach for evaluating plain language summaries adopts evaluation metrics for summarization and simplification, coupled with human evaluation \citep{jain2021summarization, ondov2022survey}. While ROUGE \citep{lin2004rouge} and BLEU \citep{sulem2018bleu} are frequently employed in PLS assessment, their efficacy is limited due to the reliance on high-quality reference summaries, which are often challenging to obtain for PLS or may not exist at all. Further, these metrics struggle to accurately identify hallucinations, especially crucial for PLS in the health domain to accurately inform health decisions \citep{wallace2021generating, pagnoni2021understanding, wang-etal-2023-automated}. Though human evaluation offers thorough assessment \cite{hardy2019highres}, the high costs and time needed impede scalability for larger datasets. While recent progress in prompt-based evaluation shows potential for assessing factuality \citep{luo2023chatgpt} and summarization quality \citep{gao2023human}, their efficacy for PLS is yet to be validated. Our work aims to fill these gaps through a systematic examination of these metrics within the PLS context.

\vspace{-1mm}
\paragraph{Robust Analysis with Synthetic Data}
Synthetic data has been widely used in NLP to evaluate metrics, for tasks such as text generation \cite{he2022blind, sai2021perturbation}, natural language inference \cite{chen2022menli, mccoy2019right}, question answering \cite{ribeiro2019red}, and reading comprehension \cite{sugawara2020assessing}. Yet, no prior work has focused on the PLS task or incorporated simplification into their benchmarks. Additionally, previous studies lack granular analyses to capture nuanced relationships between text changes and score changes. Our research endeavors to bridge these gaps by crafting perturbations that mirror real-world errors found in PLS.

\section{PLS Evaluation Criteria}
\label{criteria_specific_perturbation_design}
\begin{table}[t!]
\small
\centering
\begin{tabular}{@{}llll@{}}
\toprule
\textbf{Dataset} &  \textbf{Version}      & \textbf{Word} & \textbf{Sentence}  \\
\midrule
\textbf{CELLS}   & Abstract (\textit{src.})    & 283$_{\pm132}$           & 11$_{\pm6}$  \\
(n=6,311)        & PLS (\textit{tgt.})    & 178$_{\pm74}$           & 7$_{\pm3}$    \\
        & Candidate summary & 134$_{\pm58}$           & 5$_{\pm2}$    \\
        & GPT-simplified    & 130$_{\pm34}$            & 6$_{\pm2}$   \\
\midrule
\textbf{PLABA}   & Abstract (\textit{src.})    & 240$_{\pm95}$           & 10$_{\pm4}$  \\
(n=750)        & Adaptation (\textit{tgt.})     & 244$_{\pm95}$           & 12$_{\pm5}$  \\
\bottomrule
\end{tabular}
\caption{Diagnostic datasets statistics (mean$\pm$std).}
\label{table:cells_plaba_dataset}
\vspace{-3mm}
\end{table}

\begin{table*}
  \includegraphics[trim={12cm 3.5cm 12cm 3.15cm},clip, scale=0.36]{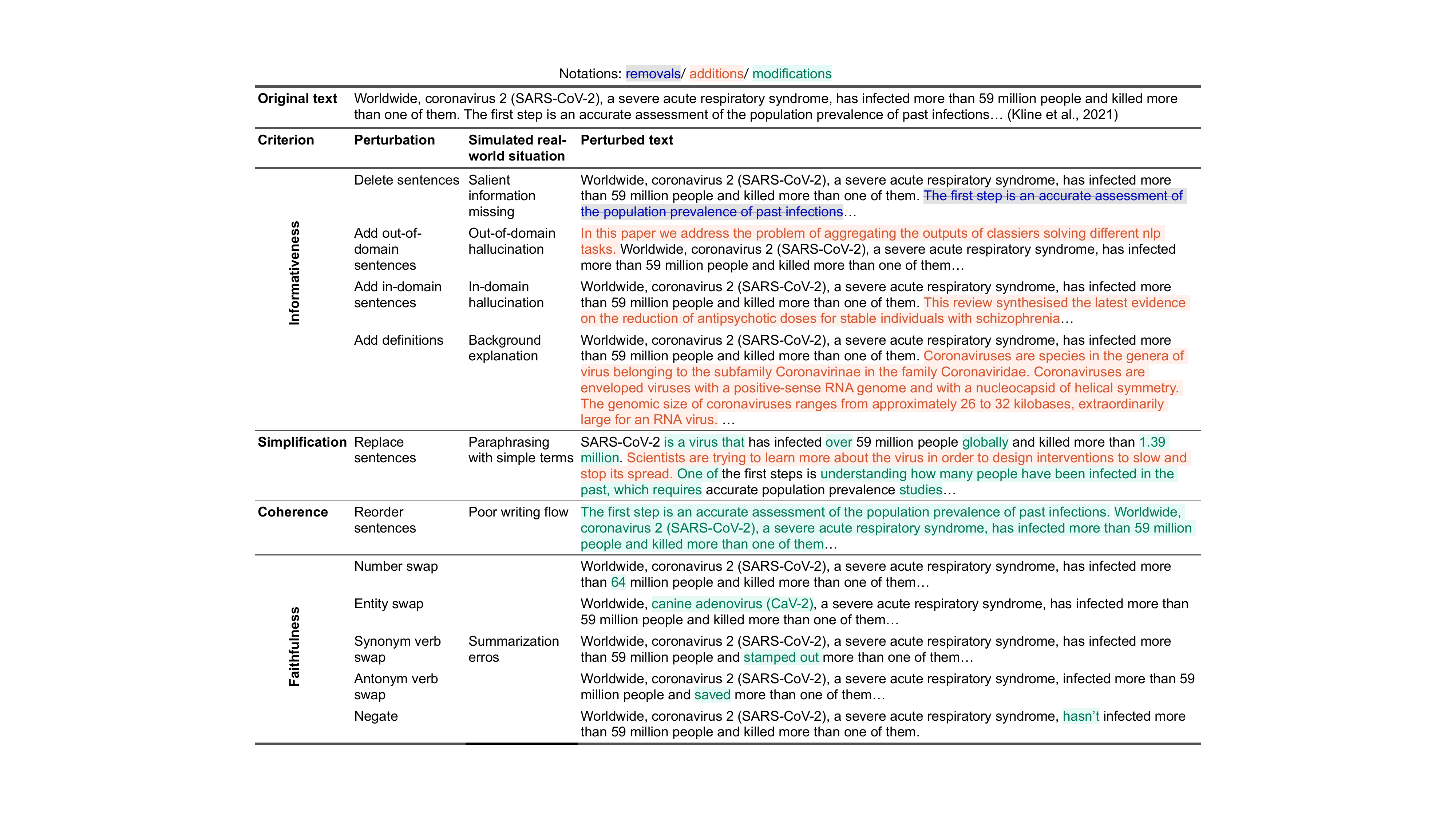}
  \vspace{-2mm}
  \caption{Example perturbations for criteria in APPLS. Original text comes from the CELLS \cite{guo2022cells}.}
  \label{table:perturbation_example}
\end{table*}

We identify four criteria that an effective PLS evaluation metric should be sensitive to based on both abstractive summarization \citep{sai2022survey, koto2022ffci} and plain language summarization paradigms \cite{pitcher2022template, ondov2022survey, stoll2022plain, jain2022survey}. 
As in \citet{gabriel2020go}, we define \textit{sensitivity} 
as being correlated in the expected direction with the amount of change in that criteria. 

\begin{itemize}[itemsep=0pt, topsep=2pt, leftmargin=10pt]
    \item \textbf{Informativeness} measures the extent to which the plain language summary covers essential information from the source text (e.g., methods, main findings) and incorporates relevant background information \cite{smith2021role, beck1991revising}. 

    \item \textbf{Simplification} describes the degree to which information is conveyed in a form that non-expert audiences can readily understand. It is distinct from informativeness because it focuses on surface-level changes (e.g., shorter sentences) but not other changes relevant to content (e.g., background explanation). 

    \item \textbf{Coherence} describes the logical arrangement of a plain language summary.

    \item \textbf{Faithfulness} denotes how well the summary aligns factually with the source text. 

\end{itemize}

\section{Constructing the \testbed Testbed}
\label{perturbation_implementation}
To assess metric sensitivity, we develop perturbations along each evaluation criteria dimension. We implement our perturbations in two large-scale PLS datasets, described in \S\ref{sec:dataset}. We follow with a discussion of how perturbations are incorporated into these datasets and our approach for managing perturbation magnitude (\S\ref{sec:perturbations}) and validating perturbation quality (\S\ref{sec:humaneval}).

\subsection{Diagnostic datasets}
\label{sec:dataset}

For our experiments, we use the CELLS \cite{guo2022cells} and PLABA \cite{attal2023dataset} datasets. CELLS \cite{guo2022cells} is a parallel corpus of scientific abstracts (\textit{source} texts) and their corresponding plain language summaries (\textit{target} texts), which are written by the abstract authors or by other domain experts. CELLS aggregates papers from 12 biomedical journals, representing a diverse set of topics and summaries, and serves as the primary dataset in our testbed. 

The PLABA \cite{attal2023dataset} dataset 
includes expert-modified biomedical abstracts, simplified to improve understanding of health-related content. 
PLABA includes sentence-level alignments, which are useful for controlled perturbations.
However, we did not select PLABA as the primary dataset due to its reliance on relatively contrived simplifications, which lack generalizability to other PLS datasets; these modifications include rule-based adjustments such as lexical simplification, shifting from passive to active voice, and segmenting long sentences.
This results in high $n$-gram overlap between sources and target summaries, which is unrealistic and does not reflect PLS in the real world.
Therefore, PLABA serves as an auxiliary dataset to CELLS, helping to address its limitations discussed in Sections \S\ref{sec:perturbations} and \S\ref{sec:humaneval}. Full results using PLABA as the diagnostic dataset are in App.~\ref{appsec:plaba_result}. 

\subsection{Applying perturbations to datasets}
\label{sec:perturbations}

\noindent Illustrative examples of all perturbations are shown in Table~\ref{table:perturbation_example}.
For the \testbed testbed, we propose and apply perturbations to a \textit{candidate summary}, which is an extractive summary constructed in the oracle setting with additional lexical variation introduced through round-trip translation \citep{ormazabal2022principled} (Illustration in App. Figure~\ref{fig:hypotehsis}).\footnote{To achieve the necessary level of control for detecting the sensitivity of automated scores to perturbation modifications, we opt to use an extractive summary instead of a language model to generate the candidate summaries.}
We do not perturb the target text directly since the resulting candidate summary would be overly similar to the target, which would be unrealistic. 

For CELLS, an extractive summary is created by selecting the set of source sentences yielding the highest ROUGE-L score when compared to the target summary, and this summary is then round-trip translated through German to derive the candidate summary. To identify the optimal extractive summary, we exhaustively evaluate all possible subsets of sentences from the source document while preserving their original order, ensuring the highest ROUGE-L score is achieved.\footnote{Why not use the extractive summary directly? Metrics like SARI expect the candidate summary to contain simplified text and exhibit degenerate behavior when used to evaluate extractive summaries directly.}
The PLABA dataset already contains sentence alignments, with sources and targets having similar lengths, so we produce the candidate summary through round-trip translation of the target alone.
Details are in App.~\ref{sec:round_trip_translation_appendix}.

We apply all perturbations to these candidate summaries as described below, where each perturbation introduces a \textit{change} (e.g., add/swap sentences) at some \textit{magnitude} (e.g., replace 50\% of sentences). Due to the high costs associated with some of our perturbations (e.g., LLM-based simplification), we restrict our testbed to the test splits of our diagnostic datasets (stats in Table~\ref{table:cells_plaba_dataset}).\footnote{The CELLS dataset contains 63k pairs; only the test split with 6.3k pairs is used for \testbed construction.} To mitigate the effects of randomness, we use two random seeds to produce all perturbations. 

\paragraph{Informativeness}
\begin{itemize}[noitemsep, topsep=1pt, leftmargin=5pt, labelindent=5pt,itemindent=-5pt]
\item[] \emph{Delete sentences}: 
To simulate the loss of relevant information, we delete sentences until a single sentence remains. The magnitude of deletion is the ratio of remaining to original sentences.
\item[] \emph{Add sentences}: We insert up to the same number of sentences as in the candidate summary. To simulate out-of-domain hallucinations, we add sentences from ACL papers \citep{bird2008acl}. For in-domain hallucinations, we add sentences from Cochrane abstracts.\footnote{\href{https://community.cochrane.org}{https://community.cochrane.org}} The magnitude of addition is the ratio of added to original sentences.
\item[] \emph{Add definitions}: Background explanations are fundamental to PLS and involve adding external content like definitions or examples \citep{guo2022cells, srikanth2020elaborative}. To simulate these, we add up to three definitions, the average number of nouns explained in CELLS \cite{guo2022cells}, i.e., 100\% perturbed adds three definitions.
\end{itemize}

\paragraph{Simplification}
\begin{itemize}[noitemsep, topsep=1pt, leftmargin=5pt, labelindent=5pt,itemindent=-5pt]
\item[] \emph{Replace sentences}: For CELLS, we first generate an LLM-simplified summary from the candidate summary. We align sentences between the candidate summary and LLM-simplified summary using the sentence alignment algorithm from \citet{guo2022cells}. We perturb the text by replacing random sentences from the candidate summary with their corresponding simplifications until full replacement is achieved.
We use GPT-4 \citep{achiam2023gpt} to generate simplifications due to its accessibility and demonstrated proficiency in text simplification \cite{Lu2023NapSSPM}. To ensure that our findings are not specific to the chosen model, we also generate simplifications and conduct experiments using GPT-3 \citep{Brown2020LanguageMA}, Llama2 \cite{touvron2023llama} and Claude.\footnote{\href{https://www.anthropic.com}{https://www.anthropic.com}. \label{claude} Access date: 12/04/2023.}
For PLABA, we perturb text by replacing source sentences with round-trip translated versions of their aligned simplified targets (no LLM is used).
\end{itemize}

\paragraph{Coherence}
\begin{itemize}[noitemsep, topsep=1pt, leftmargin=5pt, labelindent=5pt,itemindent=-5pt]
\item[] \emph{Reorder sentences}: We shuffle sentences in the candidate summary and quantify perturbation percentage in terms of the absolute difference in sentence order between the original and shuffled candidate summaries, e.g., a document with reversed sentence order would be considered 100\% perturbed. Details are in App.~\ref{app:perturbations}.
\end{itemize}

\paragraph{Faithfulness}
\begin{itemize}[itemsep=0pt, topsep=1pt, leftmargin=5pt, labelindent=5pt,itemindent=-5pt]
\item[] \emph{Number swap}: We identify numerals in the text and randomly add a number from 1 to 5 to the original numerical value.
\item[] \emph{Verb swap}: We introduce two perturbations by substituting verbs with either synonyms or antonyms. An appropriate metric should be less sensitive to synonyms but more sensitive to antonyms. 
\item[] \emph{Entity swap}: 
We replace entities using the KBIN method \cite{wright2022generating}, which replaces entity spans with related concepts in the UMLS\footnote{\href{https://www.nlm.nih.gov/research/umls/}{https://www.nlm.nih.gov/research/umls/}}
while maximizing NLI contradiction and minimizing LM perplexity. 
This results in a fluent sentence that varies from the original one. 
\item[] \emph{Negate sentences}: We negate sentences, 
and allow up to one negation per sentence. 
\end{itemize}

\vspace{2mm}
\noindent The perturbation magnitude of number, verb, and entity swaps is determined by comparing the count of altered spans to the total number of eligible spans in the candidate summary. Full perturbation means all eligible spans are swapped.

\subsection{Human validation of candidate summaries and LLM simplifications}
\label{sec:humaneval}

We validate two design decisions of \testbed that involve other models modifying text---round-trip translation (RTT) for the extractive summary and GPT-4-based simplification perturbations---by conducting human evaluation. We sample 100 pairs each of (i) extractive summaries (pre-RTT) paired with candidate summaries (post-RTT) and (ii) GPT-simplified summaries paired with candidate summaries. Annotators were asked to assess content alignment (defined as having comparable entities and relations between entities) and rate informativeness, simplification, faithfulness, and coherence on 5-point Likert scales.
Annotations were performed by two independent annotators, both with doctorates in the biological sciences, who were hired on UpWork and compensated at 21 USD/hr. Each annotator reviewed all sampled pairs for both evaluation tasks. Inter-rater agreement measured by Cohen's Kappa was 0.48, by Spearman rank correlation was 0.58, implying moderate agreement for both tasks \citep{artstein2008inter}. Details of the annotation tasks are given in App.~\ref{sec:human_evaluation_appendix}.

Human annotators affirmed that RTT text (candidate summary) retained its informativeness (98\%), faithfulness (83\%), coherence (100\%), and simplicity (96\%) compared to the extractive summary. For GPT-simplified sentences, evaluators rated its informativeness (95\%), faithfulness (95\%), coherence (98\%), and simplicity (100\%) compared to the candidate summary, with GPT-simplifications consistently rated as more simple than the candidate summary while preserving semantic content. In this context, we report the proportion of annotations equal to or better than neutral for each criterion. 
While the informativeness and faithfulness of GPT-simplified text are assessed to be very good at the passage level, the alignment algorithm used to produce sentence alignments for the simplification perturbation is imperfect and can introduce some errors. To mitigate the impact of such misalignment on the interpretation of results, we use the PLABA dataset for auxiliary diagnostics because it contains ground truth sentence-level alignments.




\begin{figure*}[tbhp!]
    \centering
    \includegraphics[trim={13cm 12cm 11cm 3cm},clip,scale=0.33]{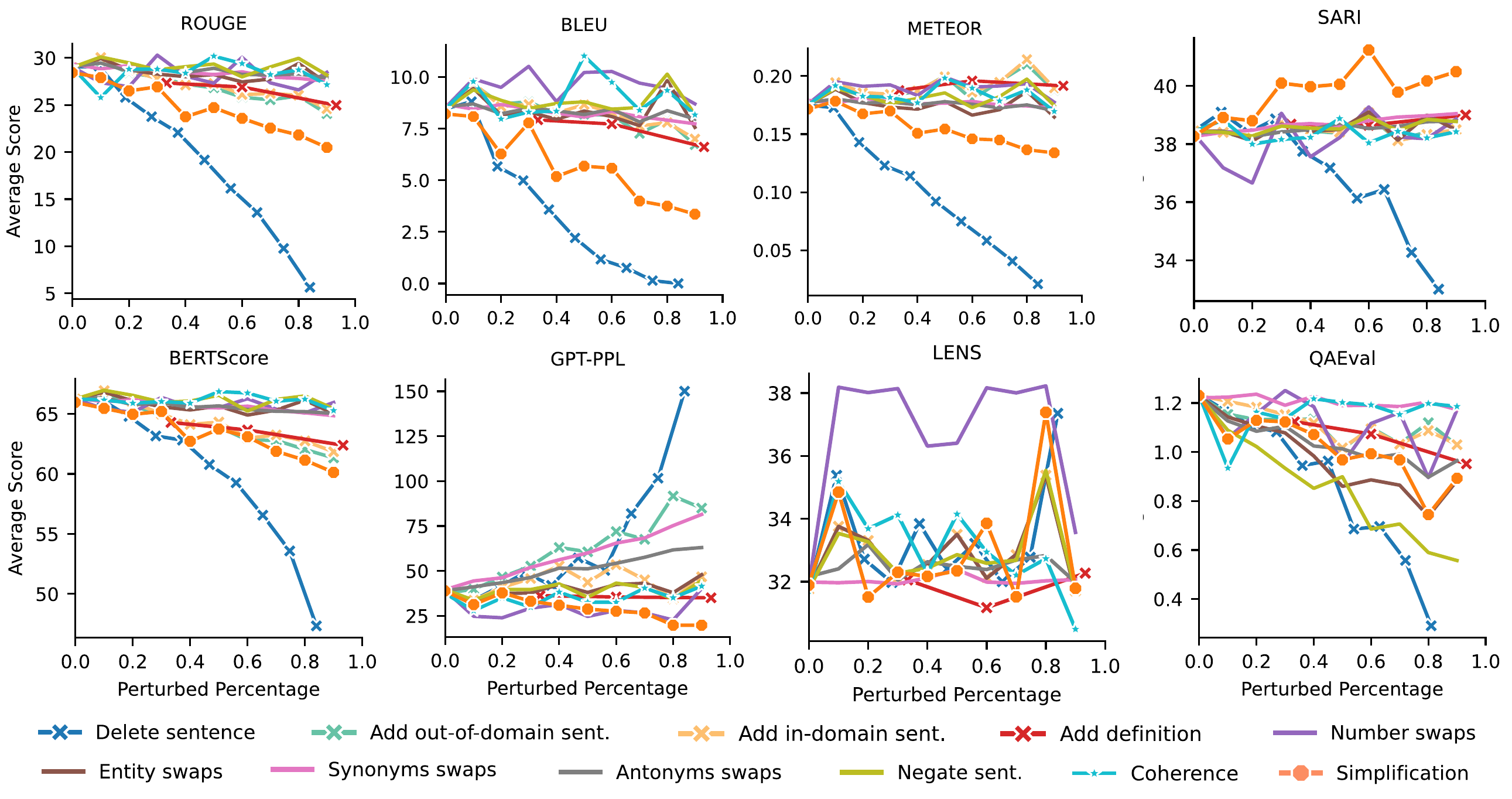}
    \caption{Average scores of existing metrics for perturbed texts in the CELLS dataset.
    Scores are averaged in 10 bins by perturbation percentage. Markers denote the defined criteria associated with that perturbation. 
    Median reported improvements in ACL'22 summarization and generation papers are ROUGE (+0.89), BLEU (+0.69), METEOR (+0.50), SARI (+1.71), BERTScore (+0.55), and PPL (-2.06). 
    }
    \label{fig:score_linechart_auto}
\end{figure*}

\section{Evaluation Metrics}
\label{existing_metrics}
Our analysis spans 8 established evaluation metrics, including the 5 most commonly reported in ACL'22 summarization/generation papers (empirical results in App.~\ref{acl_appendix}). We also assess 5 lexical features associated with text simplification (\S\ref{sec:lexical_feature_baed_metrics}) and LLM-based evaluations (\S\ref{sec:promptbased}).

\subsection{Existing automated evaluation metrics}
\label{sec:existing_automated_evaluation_metrics}

We compute the following 8 automated metrics:  

\begin{itemize}[noitemsep, topsep=1pt, leftmargin=10pt]
\item \textbf{Overlap-based metrics} measure $n$-gram overlap. We report \textit{ROUGE} (computed as the average of ROUGE-1, ROUGE-2, and ROUGE-L) \cite{lin2004rouge}, \textit{BLEU} \cite{papineni2002bleu}, \textit{METEOR} \cite{banerjee2005meteor}, and \textit{SARI} score \cite{xu2016optimizing}. 
\item \textbf{Model-based metrics} use pretrained models to evaluate text quality. We adopt \textit{GPT-PPL}, \textit{BERTScore} \cite{zhang2019bertscore}, and \textit{LENS} \cite{maddela2022lens}. 
\item \textbf{QA-based metrics} capture content quality using a question-answering approach. We report \textit{QAEval} \cite{deutsch2021towards} scores here. 
\end{itemize}

\noindent Details for all metrics are available in App.~\ref{sec:existing_automated_evaluation_metrics_app}.
All metrics assessed require target and generated texts; while SARI and LENS additionally make use of the source texts. 

\subsection{Lexical features} 
\label{sec:lexical_feature_baed_metrics}
We also assess lexical features that have been shown to be associated with text simplicity:
\begin{itemize}[noitemsep, topsep=1pt, leftmargin=10pt]
\item \textbf{Length}: Shorter sentences are easier to understand \cite{kauchak2017measuring}. We report both sentence length and paragraph length.
\item \textbf{Familiarity}: Simple text contains more common words \cite{leroy2018next}. We compute the percentage of text that is made up of the 1,000 most common English words.\footnote{\href{https://gist.github.com/deekayen/4148741}{https://gist.github.com/deekayen/4148741}}
\item \textbf{Specificity}: Specificity quantifies the level of detail in the text. We use Speciteller \cite{ko2019domain} to compute the domain agnostic specificity of terms in the paragraph.
\item \textbf{Phrase Transitions}: Conjunctions (e.g., therefore) are important for flow and can assist with comprehension \cite{kauchak2017measuring}. We report the number of conjunctions.
\item \textbf{Function Words}: Simple text contains more verbs and fewer nouns \cite{mukherjee2017role}. We report the number of verbs, nouns, adjectives, adverbs, and numbers.

\end{itemize}

\subsection{LLM prompt-based evaluations}
\label{sec:promptbased}
Prompting LLMs for text generation evaluation has been explored in recent work \cite{gao2023human, luo2023chatgpt}. We adopt the prompt template from \citet{gao2023human} to have GPT-4 (\textit{gpt-4-0613}) evaluate each candidate summary on the four PLS criteria
and to provide an overall quality score. All scores range from 0 (worst) to 100 (best). We supply definitions for each criterion in the prompt. 
We evaluate under three settings: (a) providing a single criterion in the prompt and requesting a score for that criterion; (b) providing all criteria in the prompt and requesting scores for each criterion as well as an overall score; and (c) the same setting as (b) but requiring explanations be generated alongside the provided scores. Model configurations and prompts are in App.~\ref{sec:prompt_based_evaluation_appendix}.

\section{Analysis Results}
\label{results}

\begin{figure}[t!]
    \centering
    \includegraphics[scale=0.75]{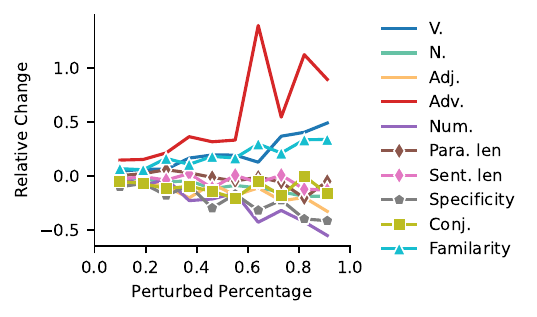}
    \vspace{-2mm}
    \caption{Relative change of each lexical feature with respect to the unperturbed state (0\%). Different markers represent lexical feature categories.}
    \label{fig:lexical_feature}
\end{figure}

\begin{figure*}[t!]
    \centering
    \includegraphics[trim={14.5cm 5cm 3cm 1cm},clip,scale=0.32]{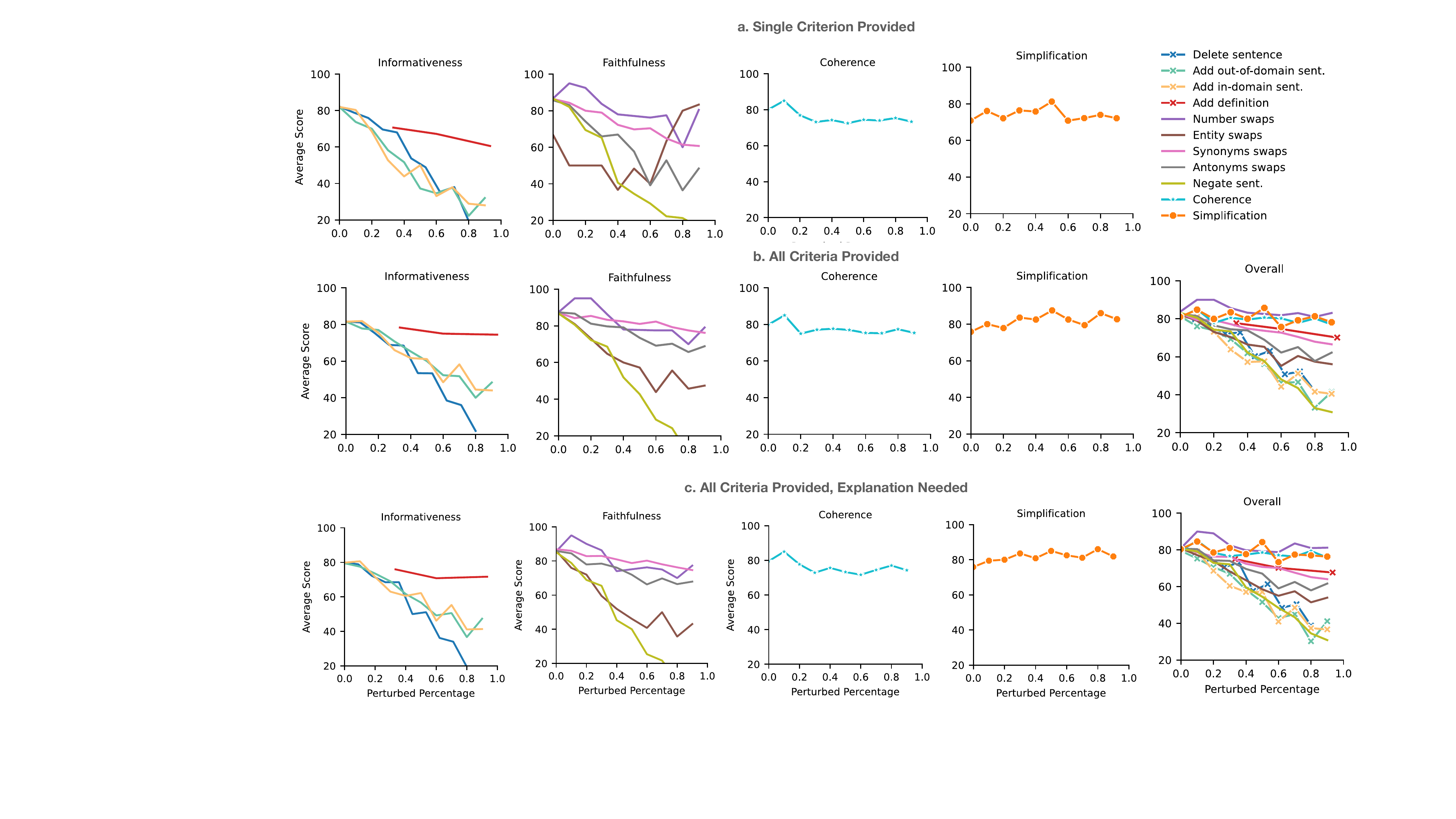}
    \caption{Prompt-based evaluation scores for four criteria - informativeness, simplification, coherence, and faithfulness - along with an overall score. (a) providing a single criterion in the prompt and requesting a score for that criterion; (b) providing all criteria in the prompt and requesting scores for each criterion as well as an overall score; and (c) the same setting as (b) but with an additional requirement for explanations of the provided scores. Notably, prompt-based scores demonstrate sensitivity to perturbations in informativeness, faithfulness, and simplification, while showing less sensitivity to changes in coherence. The three prompt settings yield similar results, with the exception that providing all criteria (setting b and c) is more sensitive to entity swaps compared to providing a single criterion (setting a).}
    \label{fig:gpt_linechart}
\end{figure*}

\setlength{\parindent}{10pt}
Automated metric responses to perturbations are in Figure~\ref{fig:score_linechart_auto}, responses of lexical features are in Figure~\ref{fig:lexical_feature}, and prompt-based evaluation results are shown in Figure~\ref{fig:gpt_linechart}. All trends are consistent across two random seeds. 

To contextualize metric performance in \testbed, we survey metric changes reported in ACL'22 papers on text generation and summarization (full results in App.~\ref{acl_appendix}). Median reported improvements for the most commonly reported metrics and SARI are: ROUGE (+0.89), BLEU (+0.69), PPL (-2.06), METEOR (+0.50), BERTScore (+0.55), and SARI (+1.71), as shown in Figure~\ref{fig:acl_score_improvement}.
\\ [-3mm]

\noindent \textbf{Aside from SARI, current metrics exhibit shortcomings in evaluating simplicity.} Metrics that are sensitive to simplification should consistently distinguish between more and less simplified text. SARI is the only automated metric among those we tested that is consistently sensitive to simplified text.  
As shown in Figure~\ref{fig:score_linechart_auto}, metrics that exhibit sensitivity to simplification perturbations are GPT-PPL (decreasing as more perturbations are introduced; lower PPL is better) and SARI score. However, in follow-up evaluations with PLABA (shown in App. Figure~\ref{fig:plaba_linechart_auto}), we see that GPT-PPL has undesirable sensitivity to text length, as found in prior work \citep{zhao2022improving}.

ROUGE, BLEU, METEOR, BERTScore, and QAEval decrease in response to the simplification perturbation. While they show consistent response relative to the degree of perturbation, 
they are nonetheless not useful for assessing text simplicity. When we report metric changes swapping sources and targets (perturbing simplified texts to increase complexity), these metrics \emph{also} decrease (App.~Figure~\ref{fig:reverse_src_tgt}), suggesting that they are sensitive to $n$-gram changes and not text simplicity.
LENS behaves erraticly with increasing simplification perturbation percentage, indicating that it is not a good metric for text simplicity.

\begin{figure*}[t!]
    \centering
    \includegraphics[trim={12cm 15cm 6cm 3cm},clip,scale=0.32]{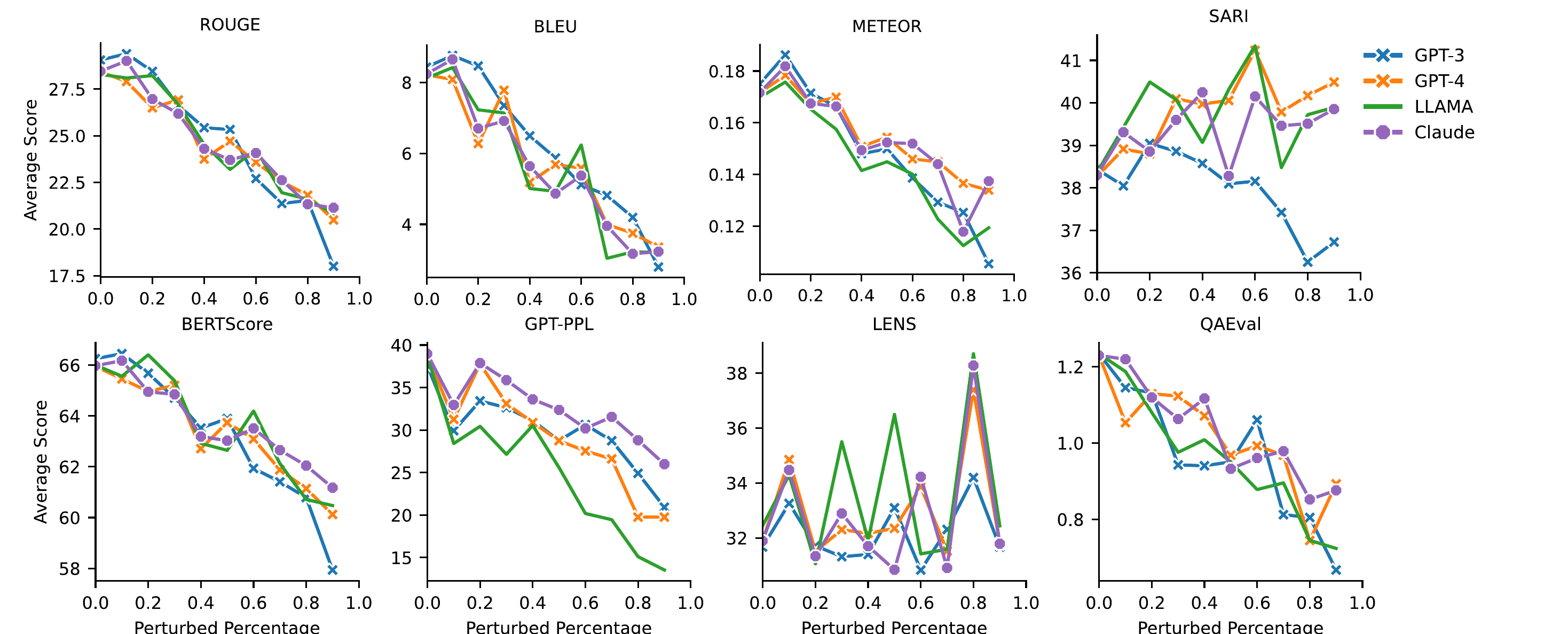}
    \caption{Variation in existing scores for simplification perturbations created by GPT-3, Llama2, and Claude on the CELLS dataset.}
    \label{fig:pomme_llama_claude}
\end{figure*}

In addition to using GPT-4 \cite{achiam2023gpt} to produced simplified text for the simplification perturbation, we also test three other LLMs: GPT-3 \citep{Brown2020LanguageMA}, Llama 2 \citep{touvron2023llama}, and Claude.\footref{claude} In Figure~\ref{fig:pomme_llama_claude}, we show metric changes to the simplification perturbation generated by all four models. Similar score changes are observed for all models (except GPT-3 for SARI score, which is an outlier), demonstrating that the simplicity perturbation in our testbed is a reasonable and consistent measure of metric response to text simplification, and that SARI is generally able to distinguish between more and less simplified text.
\\ [-3mm]

\noindent \textbf{Metrics effectively capture informativeness, coherence, and faithfulness, with room for improvement.} For informativeness, ROUGE, BLEU, BERTScore, GPT-PPL, and QAEval are sensitive to information deletion and irrelevant additions, but decrease with the addition of background explanations through keyword definitions. For coherence, BERTScore and LENS excel in detecting perturbations, largely due to their ability to assess between-sentence relationships. BERTScore, GPT-PPL, and QAEval generally perform well for faithfulness-related perturbations, although GPT-PPL and BERTScore are somewhat sensitive to synonym verb swaps instead of antonym verb swaps, which is an undesirable trait. QAEval is best at being \emph{unresponsive} to synonym verb swaps. Number swaps, however, remain undetected by all metrics. Results in Figure~\ref{fig:score_linechart_auto}.
\\ [-3mm]

\noindent \textbf{Lexical features are useful measures of text simplicity.} Figure~\ref{fig:lexical_feature}
illustrates the response of lexical features to degrees of text simplification in CELLS, confirming trends observed in previous studies \cite{kauchak2014text, leroy2018next, kauchak2017measuring, mukherjee2017role}. As simplification increases, sentence length decreases; common words and verbs increase; and nouns, adjectives, and term specificity decrease. Although prior work emphasizes the importance of conjunctions for comprehension \cite{kauchak2017measuring}, our study reveals a reduction rather than increase in conjunctions as texts become simpler. Overall, these trends demonstrate that lexical features are valuable indicators for text simplification. Results on PLABA are similar, with an inverse trend for paragraph length (App.~Figure~\ref{fig:lexical_feature_plaba}). 
\\ [-3mm]

\noindent \textbf{LLM prompt-based evaluations show promise in distinguishing between PLS criteria.} Prompt-based scores demonstrate sensitivity to perturbations in informativeness, faithfulness, and simplification, while showing less sensitivity to changes in coherence (Figure~\ref{fig:gpt_linechart}). 
While providing a single criterion, all criteria, and all criteria with an explanation mostly yield similar results, trends for simplification and some types of faithfulness perturbations are more clear and consistent when all criteria are provided.
This suggests that providing all criteria and requesting all scores simultaneously is most efficient and accurate. 

Our results also indicate that additional explanations are not essential for PLS evaluation (results for settings b and c are similar). However, further studies are required to better understand the decision-making process of the LLM, assess the validity of its explanations, and explore how the quality of explanations impacts the score. Prompts and detailed results are provided in App.~\ref{sec:prompt_based_evaluation_appendix}.


\section{Discussion \& Conclusion}

Recent advances point to the possibility of automated PLS; however, the multifaceted nature of PLS makes evaluation challenging. We introduce the first---to our knowledge---meta-evaluation testbed, \testbed, for evaluating PLS metrics. 

In \testbed, we apply controlled text perturbations to existing PLS datasets based on several criteria 
(informativeness, simplification, coherence, and faithfulness).
Using \testbed, we find that while some metrics reasonably capture informativeness, faithfulness, and coherence, SARI is uniquely sensitive to simplification perturbations, but exhibits insensitivity to other perturbations. Similar challenges are observed for QAEval, as no single metric was consistently sensitive to all perturbations across criteria. Therefore, an evaluation metric suite should be considered based on all desired criteria. From our results on \testbed, we identify the following metrics for each criterion as most promising from among those we tested: SARI for simplicity, GPT-PPL for informativeness, LENS for coherence, and QAEval for faithfulness. However, we warn that all of these automated metrics have limitations as identified in our results. Further research is necessary to identify more robust metrics for a comprehensive evaluation of PLS. 

The quickly improving performance of language models on a variety of tasks has placed greater emphasis on extrinsic human evaluations \citep{clark-etal-2021-thats} evaluating models in real-world use cases, often with end-users. However, extrinsic evaluations are time-intensive, difficult to implement correctly, and costly, making them only viable for the most promising models. Automated metrics offer a fast and low-cost method for identifying improvement trends even if they do not perfectly measure absolute improvement, or improvement at the instance level. Our selection of automated metrics grounded in criteria from the health communication literature offer a viable first step in evaluating systems. Initial automated evaluations can then be followed by extrinsic evaluations to ensure comprehensive analysis for real-world use.

Our \testbed testbed allows for extensible evaluation of PLS evaluation metrics. 
Although this study focuses on the health domain, \testbed can be adapted to other domains by changing the diagnostic dataset. Depending on the domain, evaluation criteria may need to be adjusted. For example, in legal contexts, faithfulness might be prioritized over informativeness, and additional criteria such as language specificity (e.g., avoidance of vague terminology) may be necessary.
Using our perturbation pipeline, \testbed can transform any PLS dataset into a granular meta-evaluation testbed. New perturbations can be introduced, and new evaluation metrics can also be incorporated easily into analysis.  
Our testbed lays the groundwork for further advancements in automated PLS and PLS evaluation, aiming to foster more impactful, accessible, and equitable scientific communication.

\section*{Limitations}
\label{sec:limitations}
Our perturbations use synthetic data to simulate real-world textual phenomena seen in PLS. Although our approach is informed by prior work and provides valuable insights into metric behavior, further exploration of more sophisticated methods to simulate changes in these criteria is warranted. This is especially true for aligning sentences between scientific abstracts and plain language summaries, 
as sentence-level alignment for scientific summaries is still an open problem  \citep{longeval23}.

We also acknowledge that text quality may deteriorate with synthetic perturbations in a way that affects multiple PLS criteria. However, by using synthetic data, we are benefiting from the ability to control our perturbations and extend our testbed creation framework to any dataset. It is infeasible to find naturally occurring text with the same controlled levels of each perturbation, with minimal changes to other aspects. Our aim is not to produce perfect outputs, but rather to establish a robust baseline for evaluating the performance of automated metrics for PLS evaluation. The results of our analysis complement qualitative examinations of model output conducted in other work, which further suggests that automated text generation evaluation metrics may be limited in their ability to assess generation performance of post-GPT-3 LLMs \citep{Goyal2022NewsSA}. 

We have also focused our analysis on commonly used metrics reported in prior work on simplification, summarization, and generation. Investigating the performance of metrics not included in this work, as well as the generalizability of our methods to meta-evaluation for other generative NLP tasks, is a future goal.

\section*{Acknowledgements}
This research was supported in part by the US National Library of Medicine [grant number R21LM013934] and by Azure cloud credits provided by the UW eScience Institute.


\bibliography{anthology,custom}

\clearpage
\newpage
\appendix

\section{Applying Perturbations}
\label{app:perturbations}
\begin{figure}[t!]
    \centering
    \includegraphics[trim={15cm 0cm 15cm 0cm},clip, width=0.45\textwidth]{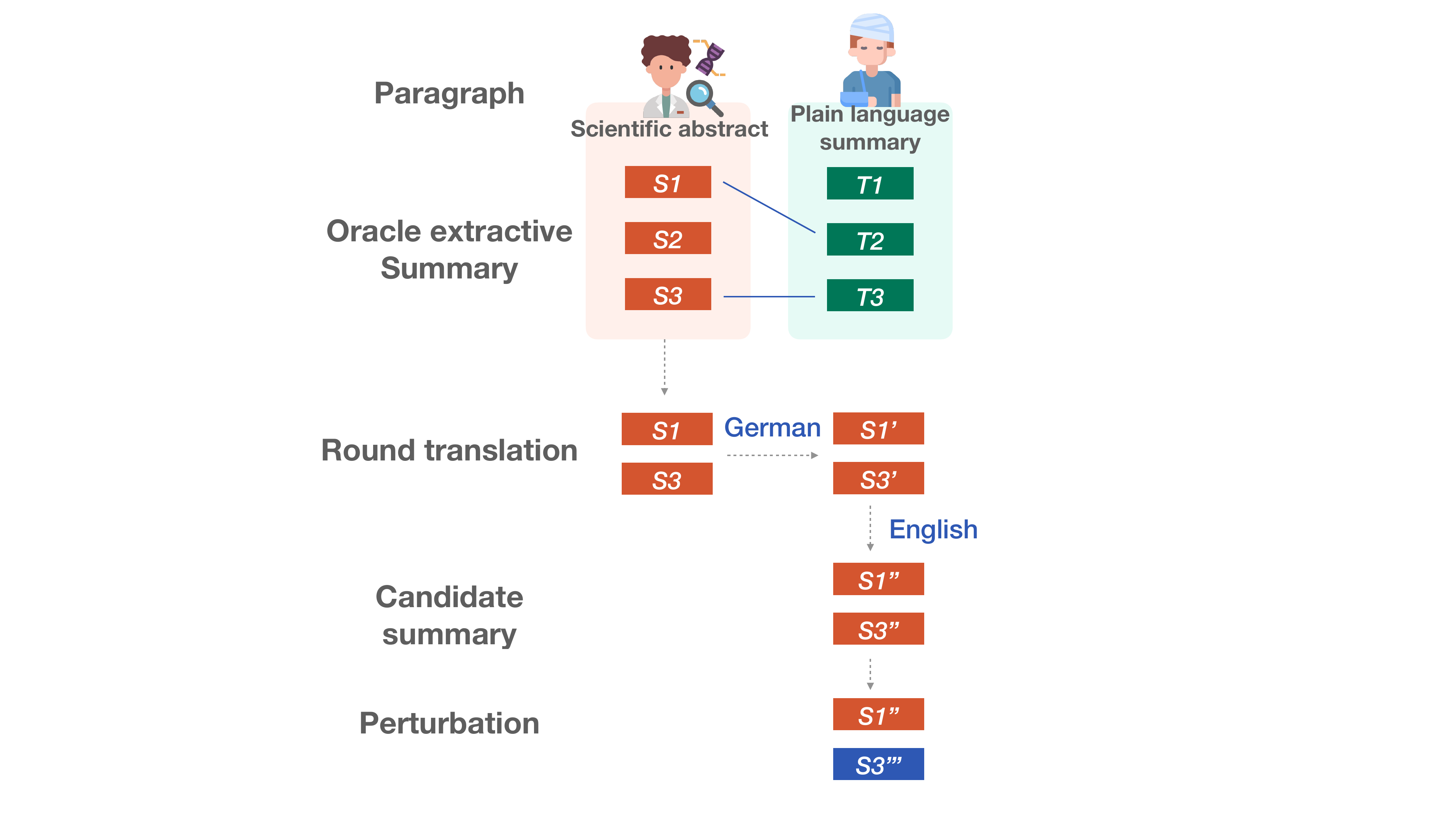}
    \caption{Process for generating the candidate summary, on which we apply all perturbation operations.}
    \label{fig:hypotehsis}
\end{figure}

Figure~\ref{fig:hypotehsis} shows how the candidate summary is extracted from the original scientific abstract. An extractive summary is identified based on high ROUGE-L with the plain language target. The extractive summary is passed through round-trip translation via German to introduce lexical variation. The resulting candidate summary forms the basis for our perturbations. 

For coherence, we count the total sentence displacement from their original positions, so for example, swapping the first and last sentences would result in a higher perturbation percentage compared to swapping the first and second sentences. We have published the perturbation code so that others can review it and deploy the testbed on different datasets.

\section{Round-trip translation for oracle extractive summary} \label{sec:round_trip_translation_appendix}

\begin{figure}[t!]
    \centering
    \includegraphics[width=0.36\textwidth]{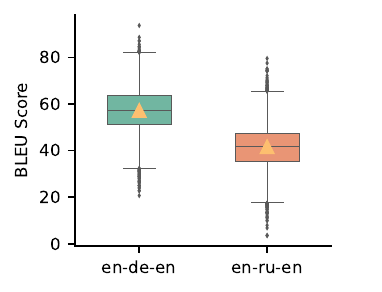}
    \caption{BLEU scores of round-trip translation for English-German-English (en-de-en) and English-Russian-English (en-ru-en) in CELLS oracle extractive summaries.}
    \label{fig:round_trip_translation}
\end{figure}
 
\begin{figure}[t!]
    \centering
    \includegraphics[width=0.36\textwidth]{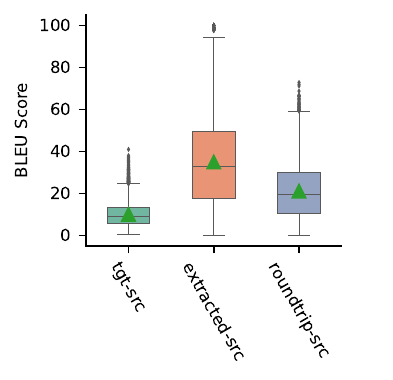}
    \caption{Comparison of BLEU scores between oracle extractive summary (extracted) and candidate summary (following roundtrip translation), using the scientific abstract (src) as the reference for BLEU calculation. }
    \label{fig:bleu_score_rtt_oracle_comparison}
\end{figure}

\begin{table*}[t!]
\small
\centering
\begin{tabular}{@{}l c c r r r r r@{}}
\toprule
 \textbf{Type} & \textbf{Unmatched} &  \textbf{Criteria} & \textbf{Str. Agree} & \textbf{Agree} & \textbf{Neutral} & \textbf{Disagree} & \textbf{Str. Disagree} \\
\midrule
\multirow{4}{*}{Round Trip Translation} & \multirow{4}{*}{1} & Simplification &  12 & 27 & 152 & 7 & 0 \\ 
& & Informativeness & 188  & 4 & 3 & 4 & 0  \\
& & Faithfulness & 155 & 6 & 4 & 20 & 14  \\
& & Coherence & 30 & 11 & 156 & 2 & 0  \\ 

\midrule

\multirow{4}{*}{GPT Simplification} &  \multirow{4}{*}{0} & Simplification  &  67 & 32 & 1 & 0 & 0 \\ 
& & Informativeness & 37 & 37 & 21 & 5 & 0  \\
& & Faithfulness & 38 & 43 & 14 & 4 & 1  \\
& & Coherence & 10 & 47 & 41 & 2 & 0  \\

\bottomrule
\end{tabular}
\caption{Counts of human evaluation ratings on each matched sentence for each criteria. For round trip translation, there are 200 ratings; for GPT simplification, there are 100 ratings. Overall, we see that round trip translation maintains strong faithfulness to the original, does not remove important information, and remains equally simple and coherent (shown by a majority of neutral ratings for the simplification and coherence criteria). For GPT simplification, we see that the simplification perturbation leads to substantially more simple text, while also maintaining faithfulness and informativeness.}
\label{human_eval_results}
\end{table*}

\begin{figure*}[t!]
  \includegraphics[trim={7cm 2cm 7cm 2cm},clip, scale=0.3]{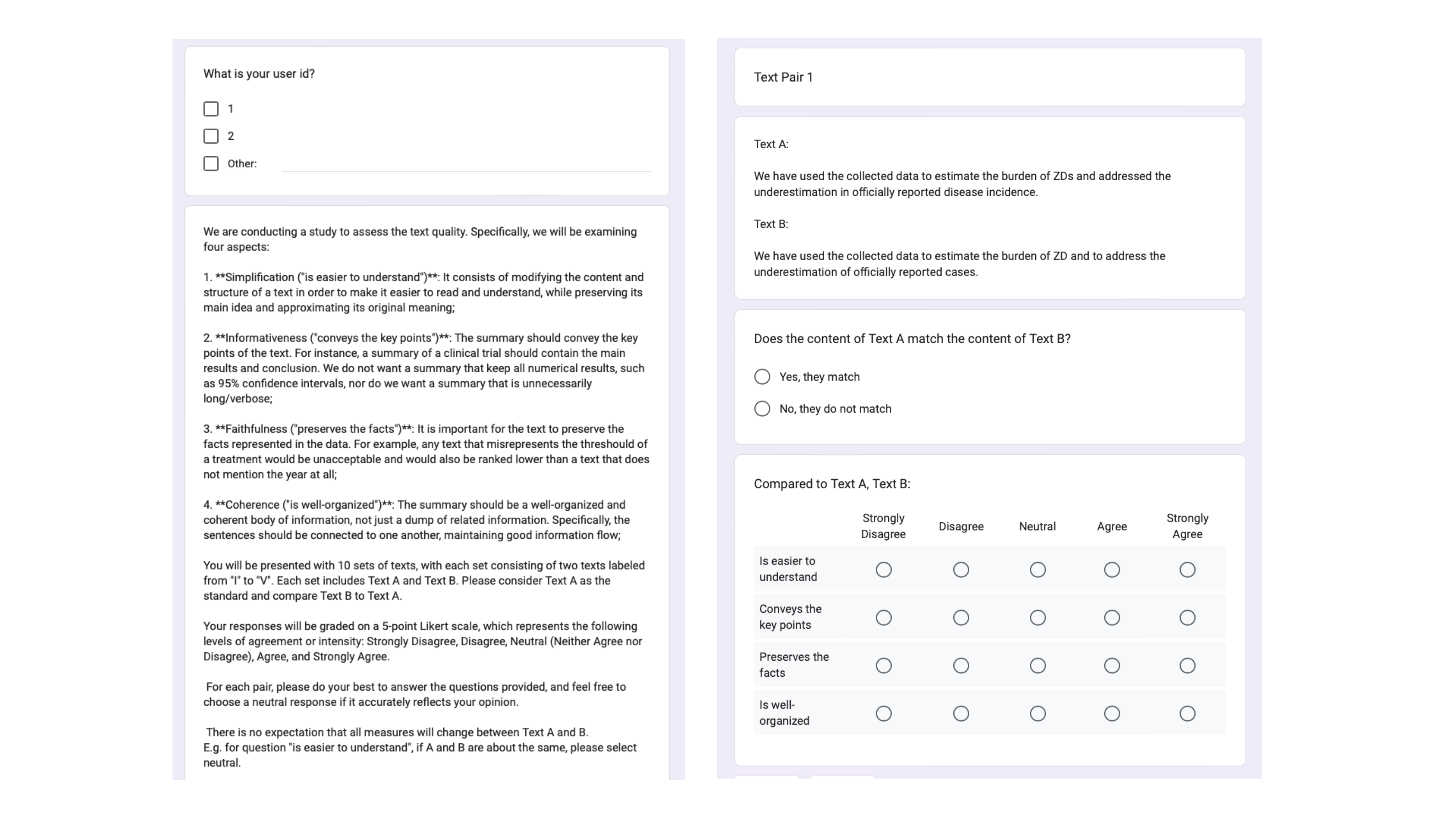}
  \caption{An example human evaluation task for assessing GPT-simplified summary quality.}
  \label{fig:human_eval_sample}
\end{figure*}

We use round-trip translation to introduce lexical variation into our oracle extractive summaries. This is important when computing metrics such as SARI, which exhibit degenerate behavior when the hypothesis is an extractive subset of the source. We examine two languages for round-trip translation: German and Russian. By employing the BLEU score as a performance metric for the round-trip generated text relative to the original source, we find that the English-German-English (en-de-en) translation sequence yields superior BLEU scores (Figure~\ref{fig:round_trip_translation}), and therefore, select the en-de-en sequence to produce the candidate summary for our testbed.

To scrutinize the introduced variation through this extractive and round-trip translation pipeline, we evaluate the BLEU score. As depicted in Figure~\ref{fig:bleu_score_rtt_oracle_comparison}, the BLEU score for the candidate summary is lower than that of the oracle extractive summary. This suggests the successful introduction of text variations. Augmented by human evaluation results in Table~\ref{human_eval_results}, with 152 out of 198 raters indicating comparable simplification levels between the candidate summary and its extractive counterparts, we conclude that our extractive and round-trip translation approach successfully introduces lexical variation in our oracle extractive summaries without altering their simplicity level.

\section{Details of human evaluation} \label{sec:human_evaluation_appendix}

To validate the quality of candidate summaries and GPT-simplified summaries, we randomly select 100 summary pairs from each corpus for human evaluation. Each pair in the candidate summary annotation task consists of an oracle extractive sentence and its respective en-de-en round-trip-translation sentence. Similarly, each pair in the GPT-simplified summary annotation task contains a chunk of the candidate summary along with its corresponding GPT-simplified chunk. 

Each pair is reviewed by two independent annotators. Annotators were hired through UpWork and have Bachelors and Doctorate degrees in the biological sciences. In the evaluation, the text pairs are labeled as Text A and Text B, without any indication that either text is generated. The annotators are first asked to assess whether the content of Text A matches the content of Text B, where a match is defined as containing the same relation tuples. If the texts match, the annotators further evaluate Text B in relation to Text A, assessing whether Text B encapsulates key points (informativeness), is more comprehensible (simplification), maintains factual integrity (faithfulness), and exhibits a well-structured layout (coherence). All facets are assessed using a 1-5 Likert scale (1-strongly disagree, 5-strongly agree). Representative questions can be found in Figure~\ref{fig:human_eval_sample}. This research activity is exempt from institutional IRB review. 


\section{Empirical Study of Evaluation Metrics Reported in ACL 2022 Publications} \label{acl_appendix}


Our study undertakes a comprehensive analysis of scores reported in the long papers of ACL 2022 to identify the most prevalently reported metrics in summarization and simplification tasks. We primarily concentrate on tasks related to generation, summarization, and simplification. Our inclusion criteria are: 1) long papers with `generat,' `summar,' or `simpl' in the title; and 2) papers that report scores for both the current model and at least one baseline model in the main text. We exclude scores from ablation studies. 

\begin{figure}[t!]
    \centering
    \includegraphics[clip,scale=0.95]{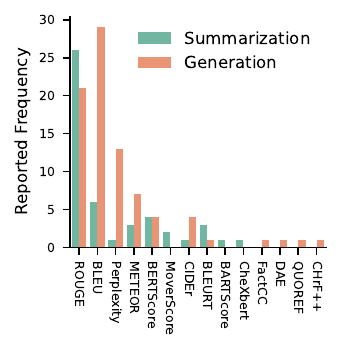}
    \caption{Most common evaluation metrics reported in ACL'22 summarization and generation long papers.}
    \label{fig:reported_score_count}
\end{figure}

\begin{figure}[t!]
    \centering
    \includegraphics[clip,scale=0.95]{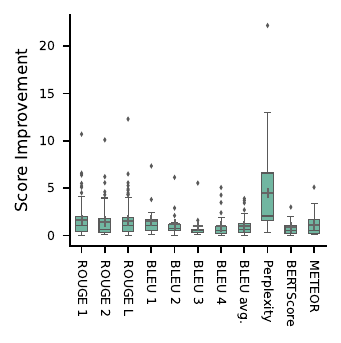}
    \caption{Distributions of reported metric improvements over baseline (absolute value) reported in ACL'22 summarization and generation long papers.}
    \label{fig:acl_score_improvement}
\end{figure}

Of the 601 long papers accepted to ACL 2022, 109 satisfy our inclusion criteria, which we categorize into 31 summarization and 78 generation papers, with no qualified papers related to simplification tasks. Considering the significance of simplification in PLS, we expanded our search to all ACL 2022 papers, including long, short, system demonstration, and findings papers. This led to the identification of 2 out of 22 papers with `simpl' in the title that reported SARI scores. As illustrated in Figure \ref{fig:reported_score_count}, the five most frequently reported automated evaluation metrics are ROUGE, BLEU, GPT-PPL, METEOR, and BERTScore.

This investigation provides insight into the current adoption of evaluation metrics in natural language generation, summarization, and simplification tasks. We observe that a majority of papers employ the same metrics across these tasks, and the reported improvements are often relatively small compared to the overall ranges for each measure. We also underscore the difficulty of interpreting changes in some of these metrics, especially model-based metrics, which lack grounding to lexical differences in text such as $n$-gram overlap.

By presenting the reported score differences from ACL papers, we hope to contextualize the metric changes observed through testing in our meta-evaluation testbed. We report the median of BERTScore values and deltas as reported in these publications, without considering the usage of different models or settings.

\section{Details on existing automated evaluation metrics}
\label{sec:existing_automated_evaluation_metrics_app}

\noindent \textbf{Overlap-based metrics} measure $n$-gram overlaps, and are popular due to their ease of use.
\begin{itemize}[noitemsep, topsep=1pt, leftmargin=10pt]
\item \textbf{ROUGE}\footnote{Implementation: \citet{fabbri2021summeval} BERTScore hash code: bert-base-uncased\_L8\_no-idf\_version = 0.3.12(hug\_trans=4.27.3).\label{summeval}} \cite{lin2004rouge} measures $n$-gram overlap between generated and reference summaries, focusing on recall. We report the average of ROUGE-1, ROUGE-2, and ROUGE-L.
\item \textbf{BLEU}\footref{summeval} \cite{papineni2002bleu} computes $n$-gram precision of generated text against reference texts, including a brevity penalty.
\item \textbf{METEOR}\footref{summeval} \cite{banerjee2005meteor} employs a relaxed matching criterion based on the F-measure, and addresses the exact match restrictions and recall consideration of BLEU.
\item \textbf{SARI}\footnote{Implementation: \citet{alva-manchego-etal-2019-easse}\label{easse}} \cite{xu2016optimizing}  is specifically designed to evaluate text simplification tasks. The score weights deleted, added, and kept $n$-grams between the source, generated, and target texts.
\end{itemize}

\noindent \textbf{Model-based metrics} use pretrained models to evaluate text quality.
\begin{itemize}[noitemsep, topsep=1pt, leftmargin=10pt]
\item \textbf{GPT-PPL},\footnote{\href{https://huggingface.co/transformers/v3.2.0/perplexity.html}{https://huggingface.co/transformers/v3.2.0/perplexity.html}} usually computed with GPT-2, measures fluency and coherence by calculating the average log probability assigned to each token by the GPT model, with lower scores indicating higher fluency and coherence.
\item \textbf{BERTScore}\footref{summeval} \cite{zhang2019bertscore} quantifies the similarity between candidate summaries and targets using contextualized embeddings from the BERT model, computing the F1-score between embeddings to capture semantic similarity beyond $n$-gram matching.
\item \textbf{LENS} \cite{maddela2022lens} employs an adaptive ranking loss to focus on targets closer to the system output in edit operations (e.g., splitting, paraphrasing, deletion).
\end{itemize}

\noindent \textbf{QA-based metrics} capture content quality using a question-answering approach. 
\begin{itemize}[noitemsep, topsep=1pt, leftmargin=10pt]
\item \textbf{QAEval} \cite{deutsch2021towards} generates question-answer pairs from the target text, then uses a learned QA model to answer these questions using the generated text. The score is computed as the proportion of questions answered correctly. We report QAEval LERC scores.
\end{itemize}


\begin{figure}[t!]
    \centering
    \includegraphics[width=0.4\textwidth]{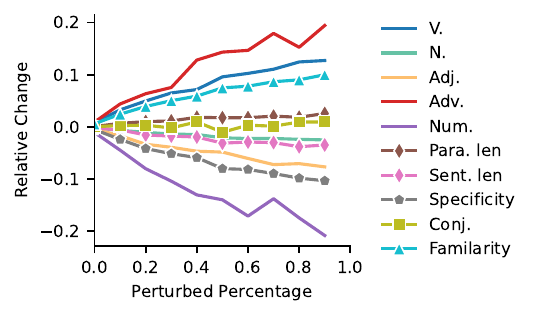}
    \caption{Relative change of each lexical feature with respect to perturbations in the PLABA dataset. Different markers represent lexical feature categories.}
    \label{fig:lexical_feature_plaba}
\end{figure}

\section{Additional experiments for existing metrics}
\begin{figure*}[t!]
    \centering
    \includegraphics[trim={13cm 15cm 12cm 13cm},clip,scale=0.37]{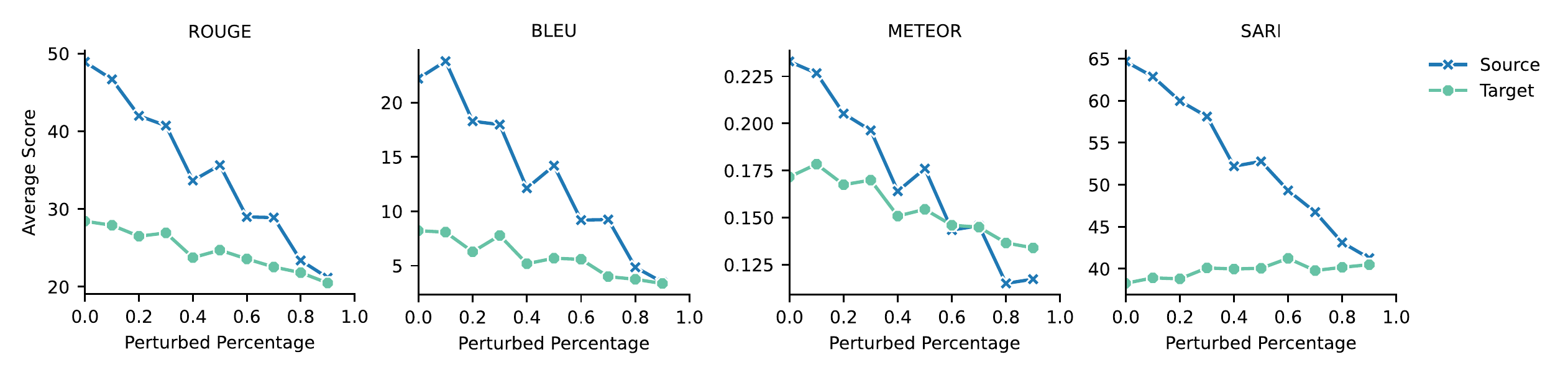}
    \caption{Average scores of ROUGE, BLEU, METEOR, and SARI scores calculated using either the source text (complex) or target text (simple) as reference for simplification perturbations on the CELLS dataset. A metric sensitive to text simplicity should move in opposing directions under these two settings. However, ROUGE, BLEU, and METEOR decrease uniformly in both settings, suggesting that they are not sensitive to text simplicity.}
    \label{fig:reverse_src_tgt}
\end{figure*}

To illustrate that existing metrics are not sensitive to text simplicity but rather to length and $n$-gram overlap, we present metric scores computed when swapping source and target for simplification perturbations (Figure~\ref{fig:reverse_src_tgt}). When target text is used as reference, we start with the candidate summary and increase perturbation percentage by swapping in simpler text, going from more complex to more simple text. When source text is used as reference, we reverse the original source and target, starting with simple text and swapping in sentences from the candidate summary, thereby moving from more simple to more complex text. A metric sensitive to text simplification should move in opposite directions in these two settings as perturbation percentage increases. However, these metric scores uniformly decrease under both settings, regardless of the reference, demonstrating that these metrics are not responsive to simplification but more so to text length and $n$-gram overlap. We do not report performance of BERTScore and QAEval under this setting due to the higher cost of computing these model based metrics.

\section{LLM Prompt-Based Evaluation}\label{sec:prompt_based_evaluation_appendix}
\begin{figure*}[tbh!]
    \centering
    \includegraphics[trim={18cm 10cm 18cm 1cm},clip,scale=0.5]{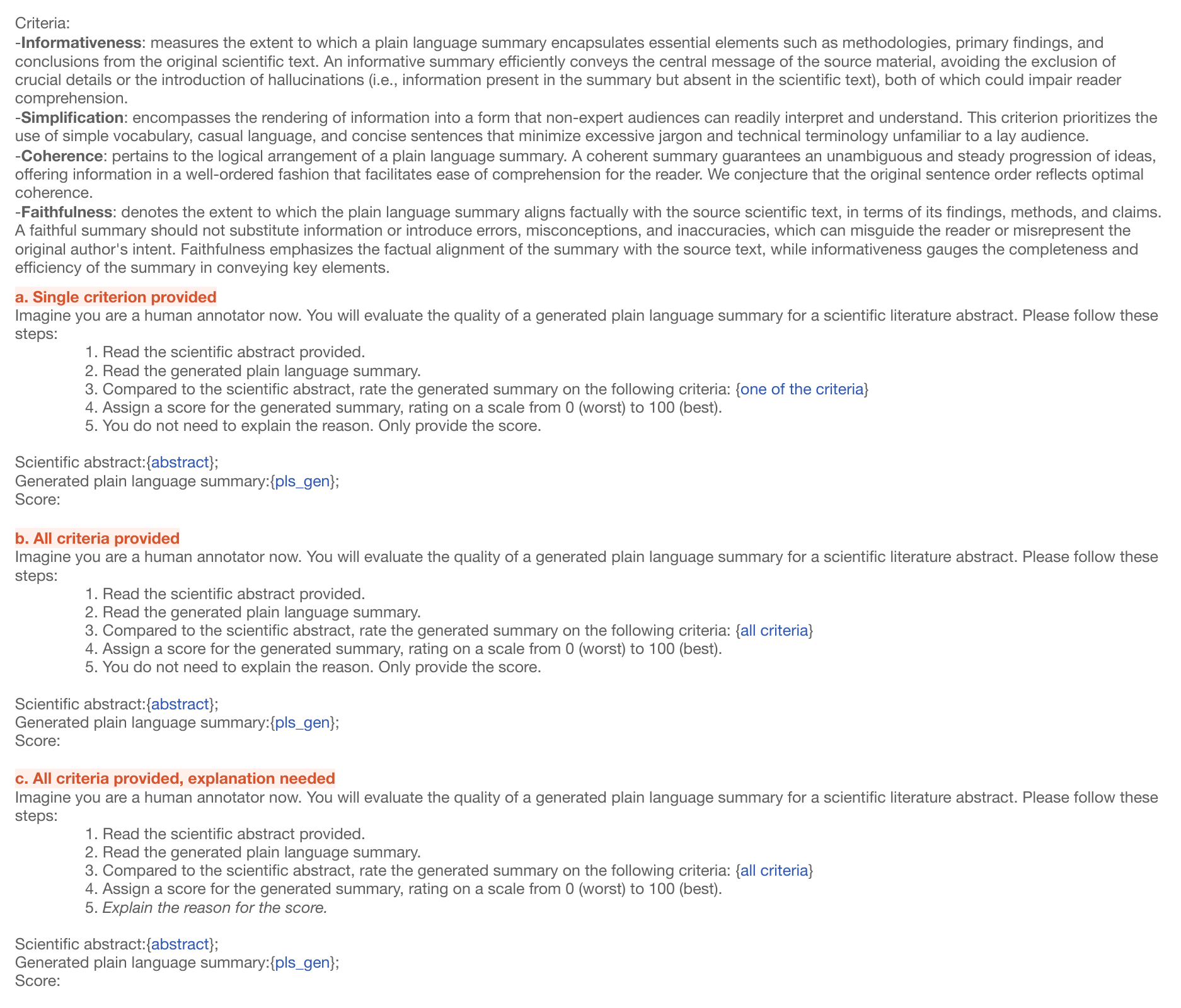}
    \caption{Prompts used for GPT-4 based evaluation: (a): single criterion provided; (b) all criteria provided; and (c) all criteria provided and explanation needed.}
    \label{fig:gpt_prompts}
\end{figure*}

We use GPT-4 for LLM evaluation. The generation process is configured with a temperature parameter of 0, a maximum length of 150, and a penalty value of 0. For each input, the top-ranked text is selected as the GPT-simplified output. 
Example prompts used for evaluation are provided in Figure~\ref{fig:gpt_prompts}. 

\begin{figure*}[t!]
    \centering
    \includegraphics[trim={10cm 10cm 8cm 1cm},clip,scale=0.32]{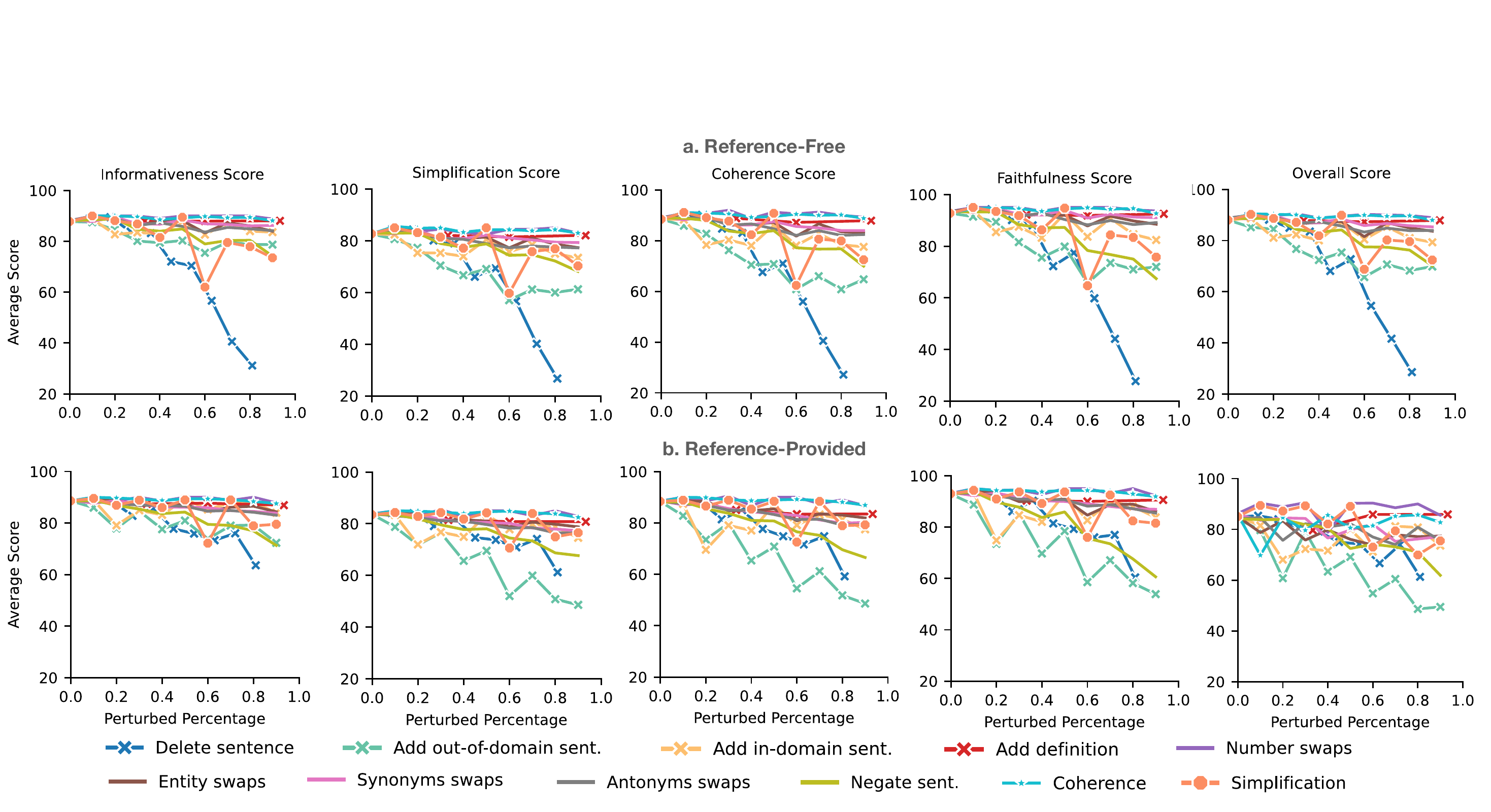}
    \caption{Prompt-based evaluation scores for four criteria - informativeness, simplification, coherence, and faithfulness - along with an overall score. (a): Reference free; (b) Reference provided.}
    \label{fig:gpt3_linechart}
\end{figure*}

\section{Additional perturbation results for PLABA}
\label{appsec:plaba_result}

\begin{figure*}[t!]
    \centering
    \includegraphics[trim={7cm 8cm 19cm 7cm},clip,scale=0.35]{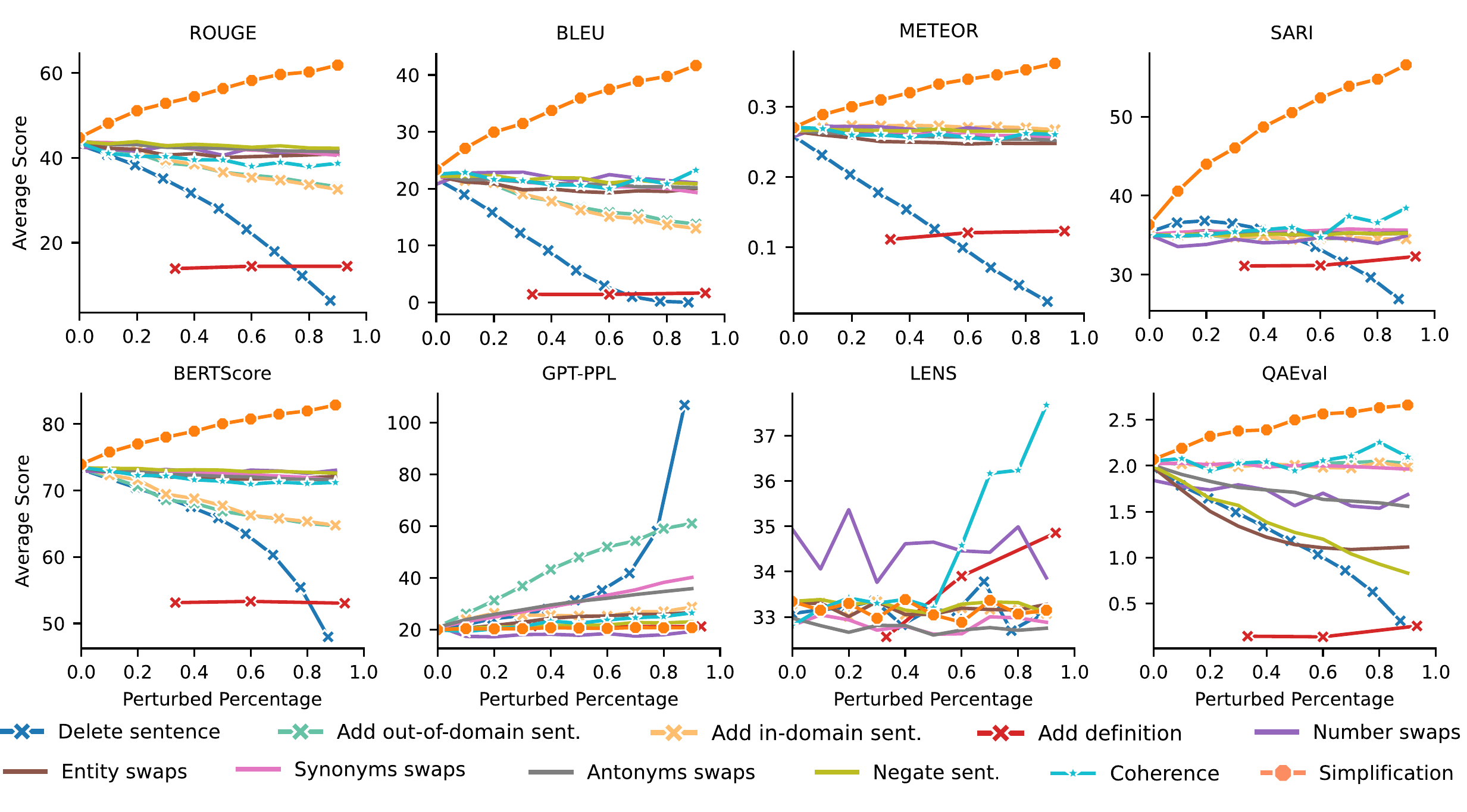}
    \caption{Average scores of existing metrics for perturbed texts in the PLABA dataset. Scores are averaged in 10 bins by perturbation percentage. Markers denote perturbations associated with our four defined criteria.}
    \label{fig:plaba_linechart_auto}
\end{figure*}

We present full perturbation results on PLABA \citep{attal2023dataset} in Figure~\ref{fig:plaba_linechart_auto}. Trends for many perturbations are in the same direction as in CELLS. While many metrics now show a desirable reversed trend to simplification (increasing), we point out that this is inconsistent performance relative to CELLS and is due to the high $n$-gram overlap between the candidate summaries and targets in this case (we perturb by replacing source sentences with round-trip translated target sentences to form the candidate summary, which only introduces minor lexical variation). Adding text, especially definitions, dramatically decreases many of these metrics due to the similar lengths of source and target texts in PLABA, again pointing to the $n$-gram and length sensitivities of most of these metrics.

The impact of simplification perturbations on lexical features in the PLABA dataset is shown in Figure~\ref{fig:lexical_feature_plaba}. Most trends are similar to CELLS, though paragraph length increases with higher perturbation percentage. In PLABA's target construction scheme, the target simplified texts length (244) are similar to the source abstracts (240).

\end{document}